\documentclass[letterpaper, 10 pt, conference]{ieeeconf}  % Comment this line out if you need a4paper

\IEEEoverridecommandlockouts                              % This command is only needed if 
\usepackage{graphics} % for pdf, bitmapped graphics files
\usepackage{epsfig} % for postscript graphics files
\usepackage{times} % assumes new font selection scheme installed
\usepackage{amsmath} % assumes amsmath package installed
\usepackage{amssymb}  % assumes amsmath package installed
\usepackage{gensymb}
\usepackage{multirow}
\usepackage{hhline}
\usepackage{booktabs}
\usepackage{float}
\usepackage{array}
\usepackage{tabularray}
\usepackage{makecell}
\usepackage[font=small,labelfont=bf]{caption}
\newcommand{\myparagraph}[1]{\vspace{0.05in}\noindent\textbf{#1}}

\usepackage{hyperref}
\hypersetup{
    colorlinks=true,
    linkcolor=blue,
    filecolor=magenta,      
    urlcolor=blue,
}

\usepackage{caption}% http://ctan.org/pkg/caption
\usepackage{cite}
\usepackage[subtle,bibbreaks=normal,paragraphs=tight,floats=normal,mathspacing=normal,wordspacing=normal,tracking=normal]{savetrees}

\title{\LARGE \bf
TEXterity: Tactile Extrinsic deXterity
\vspace{-1.5ex}
}

\author{
  \authorblockN{$^*$Antonia Bronars$^{1}$, $^*$Sangwoon Kim$^{1}$, Parag Patre$^{2}$ and Alberto Rodriguez$^{1}$} 
  \authorblockA{
     $^*$Equal Contribution, $^{1}$MIT, $^{2}$Magna International Inc.
     }
}

\begin{document}

\twocolumn[{%
\renewcommand\twocolumn[1][]{#1}%
\maketitle
\begin{center}
    \centering
    \captionsetup{type=figure}
    \vspace{-5mm}
    \includegraphics[width=.8\textwidth]{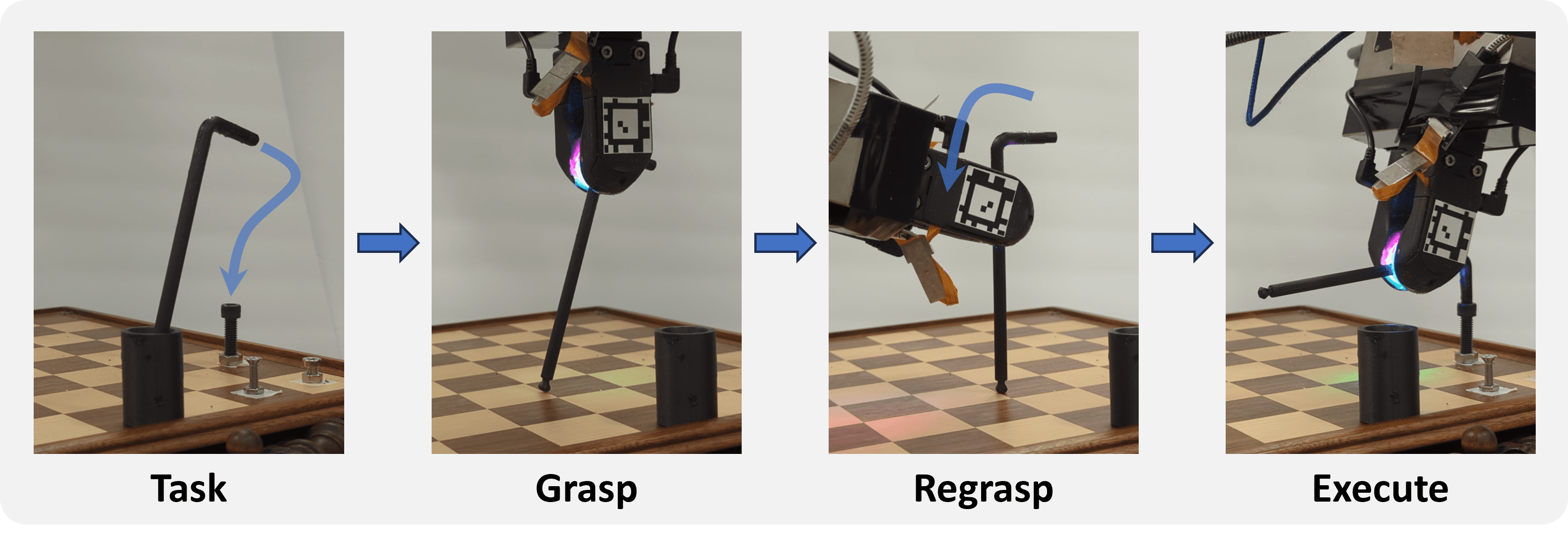}
    \captionof{figure}{An example task that requires tactile extrinsic dexterity. A proper grasp is essential when using an Allen key to apply sufficient torque while fastening a hex bolt. The proposed method utilizes tactile sensing on the robot's finger to localize the grasped object's pose and also regrasp the object in hand by pushing it against the floor - effectively leveraging extrinsic dexterity.}
    \label{fig:first}
\end{center}%
}]

\thispagestyle{empty}
\pagestyle{empty}

\begin{abstract}

We introduce a novel approach that combines tactile estimation and control for in-hand object manipulation. By integrating measurements from robot kinematics and an image-based tactile sensor, our framework estimates and tracks object pose while simultaneously generating motion plans to control the pose of a grasped object. This approach consists of a discrete pose estimator that uses the Viterbi decoding algorithm to find the most likely sequence of object poses in a coarsely discretized grid, and a continuous pose estimator-controller to refine the pose estimate and accurately manipulate the pose of the grasped object. Our method is tested on diverse objects and configurations, achieving desired manipulation objectives and outperforming single-shot methods in estimation accuracy. The proposed approach holds potential for tasks requiring precise manipulation in scenarios where visual perception is limited, laying the foundation for closed-loop behavior applications such as assembly and tool use. Please see supplementary videos for real-world demonstration at \href{https://sites.google.com/view/texterity}{https://sites.google.com/view/texterity}.

\end{abstract}

\begin{figure*}[h]
	\centering
	\includegraphics[width=0.95\linewidth]{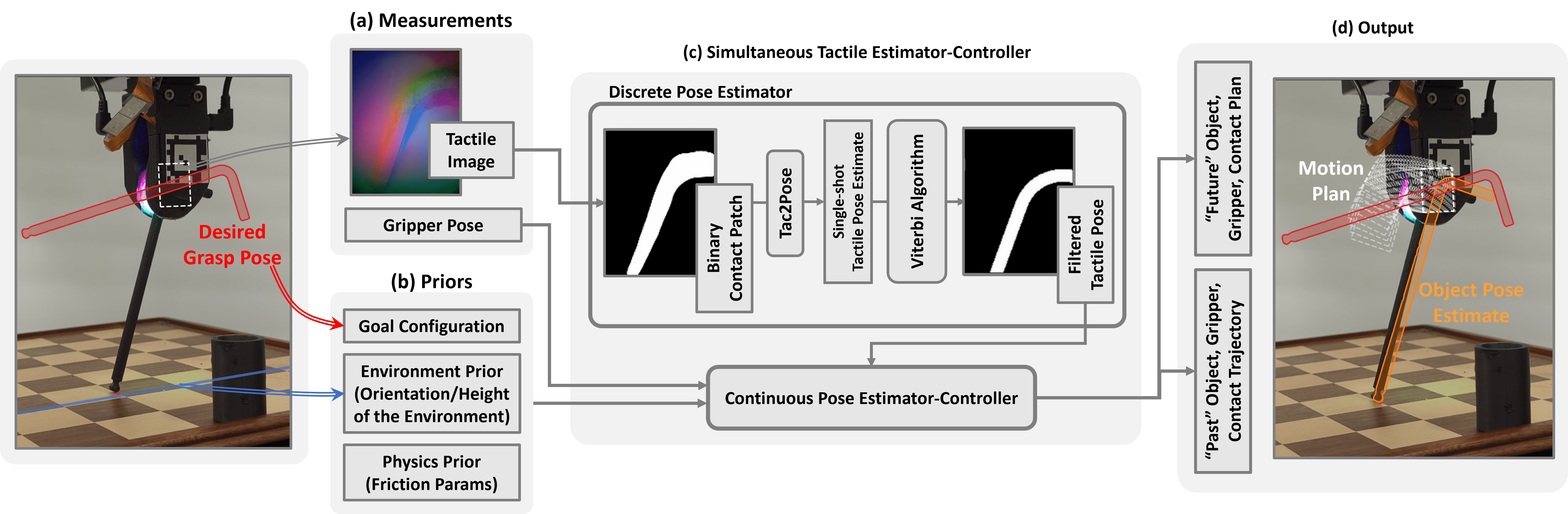}
	\caption{Overview of the Simultaneous Tactile Estimation and Control Framework.}
	\label{fig:overview}
\end{figure*}

\begin{figure*}[h]
	\centering
	\includegraphics[width=0.95\linewidth]{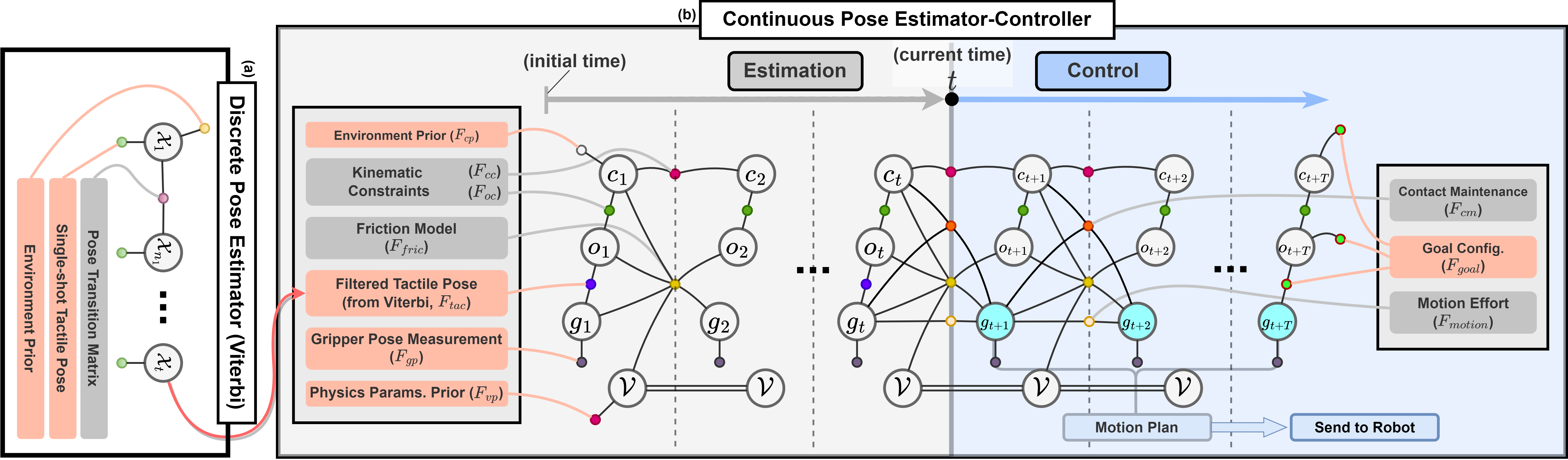}
	\caption{Graph Architecture of the Simultaneous Tactile Estimator-Controller.}
	\label{fig:graph}
    \vspace{-3mm}
\end{figure*}

%%%%%%%%%%%%%%%%%%%%%%%%%%%%%%%%%%%%%%%%%%%%%%%%%%%%%%%%%%%%%%%%%%%%%%%%%%%%%%%%
\section{INTRODUCTION}
The ability to manipulate objects within the hand is a long-standing objective in robotics for its potential to increase the workspace, speed, and capability of robotic systems. For example, the ability to change the grasp on an object can improve grasp stability and functionality, or prevent collisions and kinematic singularities. In-hand manipulation is challenging from the perspectives of state estimation, planning, and control: firstly, once the object is enveloped by the grasp, it becomes difficult to perceive with external vision systems; secondly, the hybrid dynamics of contact-rich tasks are difficult to predict \cite{bauza2018data} and optimize over \cite{mordatch2012contact}.

Existing work on in-hand manipulation emphasizes the problem of sequencing contact modes, and can be broken down into two prevailing methodologies. One line of work relies on simple object geometries and exact models of contact dynamics to plan using traditional optimization-based approaches \cite{sundaralingam2018geometric, mordatch2012contact, hou2018fast, shi2017dynamic, sundaralingam2019relaxed}, while the other leverages model-free reinforcement learning to learn policies directly \cite{chen2022system, rajeswaran2017learning, chen2023visual, handa2023dextreme, andrychowicz2020learning, huang2021generalization}. Much less consideration has been given to the challenge of precisely controlling such behaviors, despite the fact that tasks like connector insertion or screwing in a small bolt require a high precision. 

Tactile feedback is a promising modality to enable precise control of in-hand manipulation. Image-based tactile sensing \cite{yuan2017gelsight, lambeta2020digit, taylor2022gelslim} has gained traction in recent years for its ability to provide high-resolution information directly at the contact interface. Image-based tactile sensors have been used for pose estimation \cite{bauza2022tac2pose}, object retrieval \cite{pai2023tactofind}, and texture recognition \cite{luo2018vitac}. They have also been used to estimate the location of contacts with the environment \cite{kim2022active, ma2021extrinsic, higuera2023neural}, to supervise insertion \cite{dong2021tactile}, and to guide the manipulation of objects like boxes \cite{hogan2020tactile}, tools \cite{shirai2023tactile}, cable \cite{she2021cable}, and cloth \cite{sunil2023visuotactile}.

We study the problem of precisely controlling in-hand sliding regrasps by pushing against an external surface, i.e. extrinsic dexterity \cite{dafle2014extrinsic}, supervised only by robot proprioception and tactile sensing. Our framework is compatible with arbitrary, but known, object geometries and succeeds even when the contact parameters are known only approximately.

This work builds upon previous research efforts. First, \textit{Tac2Pose} \cite{bauza2022tac2pose} estimates the relative gripper/object pose using tactile sensing, but lacks control capabilities. Second, \textit{Simultaneous Tactile Estimation and Control of Extrinsic Contact} \cite{kim2023simultaneous} estimates and controls extrinsic contact states between the object and its environment, but has no understanding of the object's pose and therefore has limited ability to reason over global re-configuration. Our approach combines the strengths of these two frameworks into a single system. As a result, our method estimates the object's pose and its associated contact configurations and simultaneously controls them. By merging these methodologies, we aim to provide a holistic solution for precisely controlling general planar in-hand manipulation.

\section{RELATED WORK} \label{section:relatedwork}
\myparagraph{Tactile Estimation and Control.} Image-based tactile sensors are particularly useful for high-accuracy pose estimation, because they provide high-resolution information about the object geometry throughout manipulation. They have been successfully used to track object drift from a known initial pose \cite{sodhi2021learning, sodhi2022patchgraph}, build a tactile map and localize the object within it \cite{zhao2023fingerslam, suresh2021tactile, bauza2019tactile}, and estimate the pose of small parts from a single tactile image \cite{li2014localization}. Because touch provides only local information about the object geometry, most tactile images are inherently ambiguous \cite{bauza2022tac2pose}. Some work has combined touch with vision \cite{bauza2023simple, dikhale2022visuotactile, izatt2017tracking, anzai2020deep} to resolve such ambiguity. Our approach is most similar to a line of work which regress distributions over possible object pose from a single tactile image \cite{bauza2022tac2pose, suresh2023midastouch, kelestemur2022tactile}, then fuses information over streams of tactile images using particle \cite{suresh2023midastouch} or histogram \cite{kelestemur2022tactile} filters. \cite{suresh2023midastouch} tackles the estimation, but not control, problem, assuming that the object is rigidly fixed in place while a human operator slides a tactile sensor along the object surface. Similarly, \cite{kelestemur2022tactile} also assumes the object is fixed in place, while the robot plans and executes a series of grasp and release maneuvers to localize the object. Our work, on the other hand, tackles the more challenging problem of estimating and controlling the pose of an object sliding within the grasp while not rigidly attached to a fixture.

\myparagraph{In-Hand Manipulation.} In-hand manipulation is most commonly achieved with dexterous hands or by leveraging the surrounding environment (extrinsic dexterity \cite{dafle2014extrinsic}). One line of prior work formulates the problem as an optimization over exact models of the hand/object dynamics \cite{sundaralingam2018geometric, mordatch2012contact, hou2018fast, shi2017dynamic, sundaralingam2019relaxed, hou2020manipulation}, but only for simple objects and generally only in simulation \cite{sundaralingam2018geometric, mordatch2012contact}, or by relying on accurate knowledge of physical parameters to execute plans precisely in open loop \cite{hou2018fast}.
Another line of prior work focuses on modeling the mechanics of contact itself in a way that is useful for planning and control, either analytically \cite{shi2020hand, chavan2018hand, chavan2015prehensile} or with neural networks \cite{nagabandi2020deep, kumar2016optimal}.

Some work has avoided the challenges of modeling contact altogether, instead relying on model-free reinforcement to directly learn a policy for arbitrary geometries. Many of these policies have been tested in simulation only \cite{chen2022system, rajeswaran2017learning}, or operate on vision data \cite{chen2023visual, handa2023dextreme, andrychowicz2020learning, huang2021generalization}. They, however, suffer from a lack of precision. As an example, \cite{chen2023visual} reports 45$\%$ success on held out objects, and 81$\%$ success on training objects, where success is defined as a reorientation attempt with less than 0.4 rad (22.9\degree) of error, underscoring the challenge of precise reorientation for arbitrary objects.

There have also been a number of works leveraging tactile sensing for in-hand manipulation. \cite{she2021cable}, \cite{sunil2023visuotactile}, and \cite{tian2019manipulation} use image-based tactile sensors to supervise sliding on cables, cloth, and marbles, respectively. \cite{shirai2023tactile} detects and corrects for undesired slip during tool manipulation, while \cite{lepert2023hand} learns a policy that trades off between tactile exploration and execution to succeed at insertion tasks. Several works rely on proprioception \cite{pitz2023dextrous} or low-resolution tactile sensing like binary contact \cite{yin2023rotating}, pressure \cite{van2015learning}, or F/T \cite{sievers2022learning} sensors to coarsely reorient objects within the hand. State estimation from such sensors is challenging and imprecise, however, leading to policies that accrue large errors.

We consider the complementary problem of planning and controlling over a known contact mode (in-hand sliding by pushing the object against an external surface), where the object geometry is arbitrary but known. We leverage a simple model of the mechanics of sliding and supervise the behavior with high-resolution tactile sensing, in order to achieve precise in-hand manipulation. By emphasizing the simultaneous estimation and control for a realistic in-hand manipulation scenario, this work addresses a gap in the existing literature and paves the way for executing precise dexterous manipulation on real systems.

\begin{figure}[t]
	\centering
	\includegraphics[width=1\linewidth]{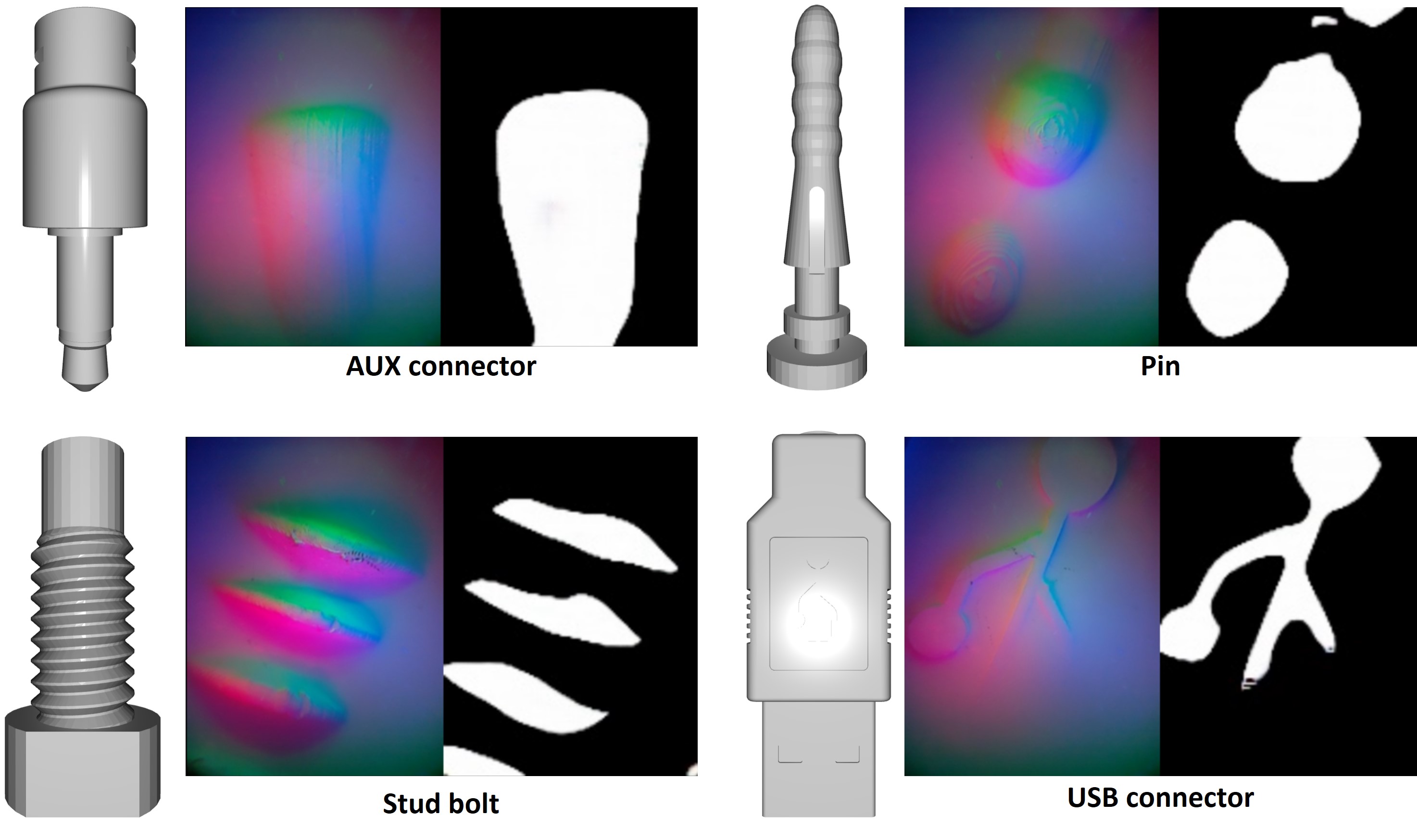}
	\caption{Test objects with example tactile images and contact patch reconstruction.}
	\label{fig:objects}
    \vspace{-5mm}
\end{figure}

\section{METHOD}

\subsection{Problem Formulation}
\label{formulation}

We address the task of manipulating objects in-hand from unknown initial grasps to achieve desired configurations by pushing against the environment. The target configurations encompass a range of manipulation objectives:

\begin{itemize}
    \item Changing the grasp pose (i.e., relative rotation/translation between the gripper and the object)
    \item Changing the orientation of the object in the world frame (i.e., pivoting against the environment)
    \item Changing the location of the extrinsic contact point (i.e., sliding against the environment)
\end{itemize}
A wide variety of target configurations with respect to the gripper and the world frames can be specified via a combination of the above objectives.

We make several assumptions to model this problem:

\begin{itemize}
    \item Grasped objects are rigid with known 3D models.
    \item The environment is flat, with a known orientation and height.
    \item Contact between the grasped objects and the environment occurs at a single point.
    \item Grasp sliding is allowed in the plane of the gripper finger surface.
\end{itemize}

\subsection{Overview} \label{subsection:overview}

Fig. \ref{fig:first} illustrates our approach through an example task: using an Allen key to apply sufficient torque while fastening a hex bolt. Adjusting the grasp through in-hand manipulation is necessary to increase the torque arm and prevent the robot from hitting its motion limit during the screwing.

Fig. \ref{fig:overview} provides an overview of the framework of our approach. The system gathers measurements from both the robot and the sensor (Fig.\ref{fig:overview}a). Robot proprioception provides the gripper's pose, while the GelSlim 3.0 sensor \cite{taylor2022gelslim} provides observation of the contact interface between the gripper finger and the object in the form of an RGB tactile image. The framework also takes as input the desired goal configuration and estimation priors (Fig.\ref{fig:overview}b):

\begin{itemize}
    \item \textbf{Desired Goal Configuration}: A combination of the manipulation objectives discussed in Section \ref{formulation}.
    \item \textbf{Physics Parameter Priors}: The friction parameters at both the intrinsic contact (gripper/object) and the extrinsic contact (object/environment). These priors do not need to be accurate and are manually specified based on physical intuition.
    \item \textbf{Environment Priors}: The orientation and height of the environment.
\end{itemize}

Utilizing these inputs, our \textbf{simultaneous tactile estimator-controller} (Fig.\ref{fig:overview}c) calculates pose estimates for the object, along with a motion plan to achieve the manipulation objectives  (Fig.\ref{fig:overview}d). This updated motion plan guides the robot's motion. The framework comprises two main components: \textbf{discrete pose estimator} and \textbf{continuous pose estimator-controller}, which are described in the next subsections.

\subsection{Discrete Pose Estimator} \label{subsection:discrete}

The discrete pose estimator computes a probability distribution within a discretized grid of relative gripper/object poses. We first reconstruct a binary mask over the region of contact from raw RGB tactile images using a pixel-to-pixel convolutional neural network (CNN) model as described in \cite{bauza2022tac2pose}. Subsequently, the binary mask is channeled into the Tac2Pose estimator \cite{bauza2022tac2pose}, which generates a distribution over possible object poses from a single contact mask.

We then merge the stream of tactile information with the environment prior using the Viterbi algorithm \cite{forney1973viterbi}, yielding a filtered probability distribution of the relative object pose. We discretize the pose space with 5mm of translational resolution and 10\degree of rotational resolution. The discretized state space consists of 5k-9k poses, depending on the object size. The inference step takes 2-6 seconds per iteration using PGMax \cite{zhou2022pgmax}, yielding a slow and coarse but global object pose signal. 

Fig. \ref{fig:graph}a provides insight into the architecture of the Viterbi algorithm, where $\mathcal{X} \in SE(2)$ represents the relative pose between the gripper and the object. At the initial timestep, the environment prior is introduced. Given our prior knowledge of the environment's orientation and height, we can, for each discrete relative object pose within the grid, ascertain which point of the object would be in closest proximity to the environment and compute the corresponding distance. The integration of the environment prior involves the multiplication of a Gaussian function over these distances. In simpler terms, it assigns higher probabilities to the relative poses that are predicted to be closer to the environment.

At each subsequent time step, we incorporate the single-shot tactile pose estimation distribution. Additionally, the transition probabilities impose constraints on tactile observations between consecutive time steps, including:
\begin{itemize}
    \item The pose can transition only to neighboring poses on the pose grid.
    \item The height of the closest point to the environment remains consistent across time steps due to the flat nature of the environment. This consistency is enforced through the multiplication of a Gaussian function that factors in the height difference.
\end{itemize}
Together, they encode the assumption that the object slides continuously within the grasp. This enables the discrete pose estimator to compute and filter the distribution of relative gripper/object poses, taking into account tactile information, robot proprioception, and environmental priors.

\begin{figure*}[h]
	\centering
	\includegraphics[width=0.93\linewidth]{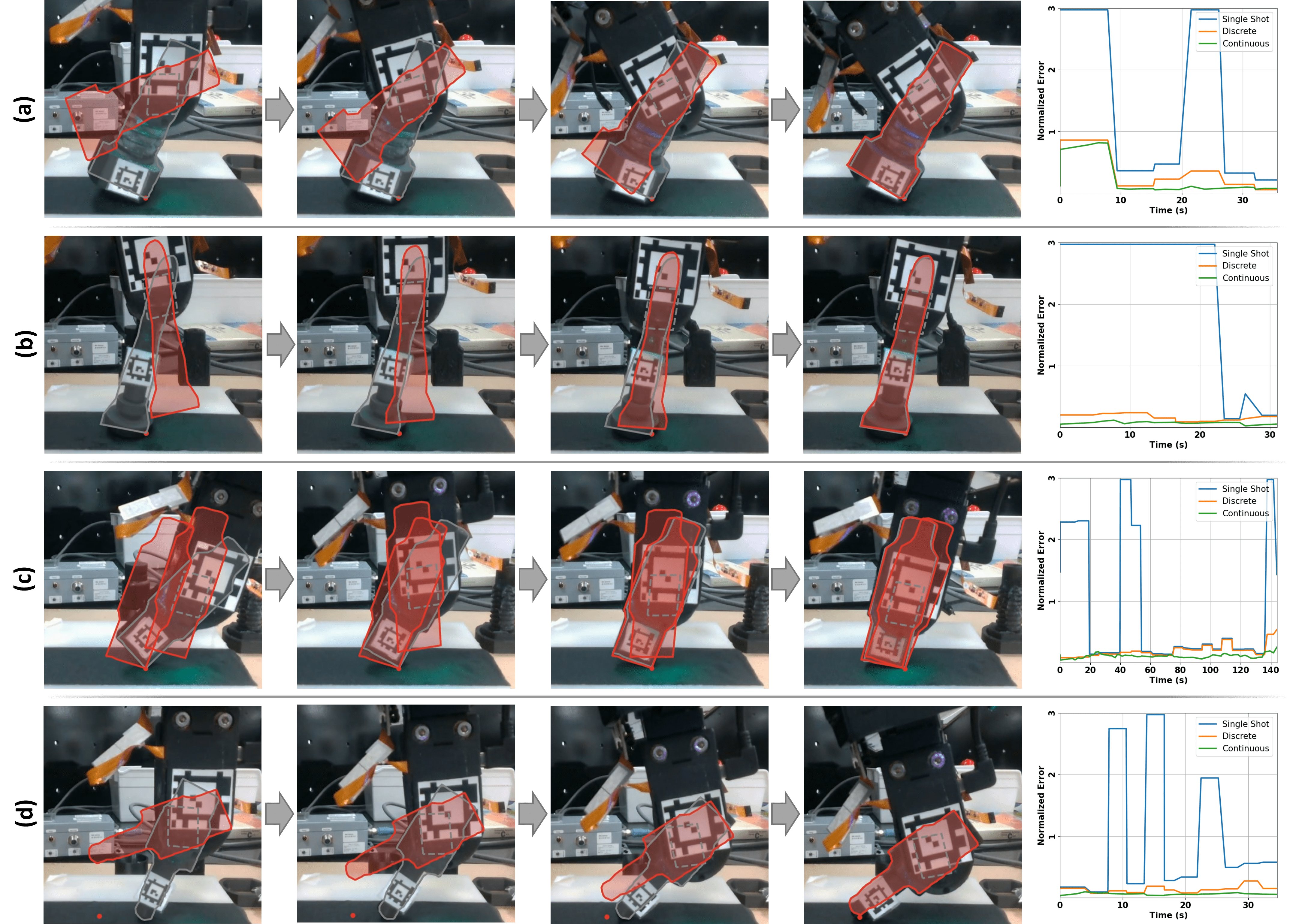}
	\caption{Demonstrations of four types of goal configurations: (a) Relative Orientation + Stationary Extrinsic Contact, (b) Relative Orientation/Translation + Stationary Extrinsic Contact, (c) Relative Orientation + Global Orientation + Stationary Extrinsic Contact, and (d) Relative Orientation + Sliding Extrinsic Contact. The right column depicts normalized estimation accuracy for the proposed method and ablation models.}
	\label{fig:samples}
    \vspace{-2mm}
\end{figure*}

\subsection{Continuous Pose Estimator-Controller} \label{subsection:continuous}

The continuous pose estimator-controller serves a dual purpose: it takes as input the filtered probability distribution of relative gripper/object poses and outputs a higher resolution pose estimate and an iteratively updated motion plan in a receding horizon fashion. The Incremental Smoothing and Mapping (iSAM) algorithm \cite{kaess2012isam2}, which is based on the factor graph model \cite{dellaert2012factor, dellaert2017factor}, serves as the computational backbone of our estimator-controller. We leverage its graph-based flexible formulation to combine estimation and control objectives as part of one single optimization problem.

The factor graph architecture of the continuous pose estimator-controller is illuminated in Fig. \ref{fig:graph}b. Noteworthy variables include $g_t$, $o_t$, and $c_t$, each frames in $SE(2)$, representing the gripper pose, object pose, and contact position, respectively. The orientation of $c_t$ is fixed and aligned with the normal direction of the environment. Additionally, $\mathcal{V}$ represents the set of physics parameters:
\begin{itemize}
    \item Translational-to-rotational friction ratio at the grasp: $F_{max}/M_{max}$, where $F_{max}$ and $M_{max}$ are the maximum pure force and torque that it can endure before sliding.
    \item Friction coefficient at the extrinsic contact between the object and the environment: $\mu_{max}$.
\end{itemize}

The framework closely resembles that of earlier work \cite{kim2023simultaneous}, where further explanation of the iSAM implementation can be found. We encourage readers to refer to this seminal work for a description of the factors that are directly borrowed from \cite{kim2023simultaneous} ($F_{cc}, F_{oc}, F_{gp}, F_{motion}$).

The continuous estimator-controller comprises two main sections: the left segment, spanning from the initial time to the current moment $t$, is dedicated to the \textbf{estimation} of the object's pose. The right segment, covering the time from $t$ to the control horizon $t+T$, is responsible for devising a motion plan to \textbf{control} the system and achieving the manipulation objectives. In the following sections, we define each new factor. The arguments of each factor definition are the variables, priors, and observations that the factor depends on. The right-hand side specifies the quantity we are trying to optimize.

\myparagraph{Estimation.} Within the estimation segment, the factor graph takes filtered pose estimations from the discrete pose estimator:
%and gripper pose measurements ($F_{gp}$).
\begin{align}
    &F_{tac}(g_i,o_i;\mathcal{X}_{i,\text{MAP}}) = \mathcal{X}_{i,\text{MAP}}^{-1} (g_i^{-1}o_i),
\end{align} 
where $\mathcal{X}_{i,\text{MAP}}$ denotes the filtered maximum a posteriori (MAP) discrete relative pose, and $(g_i^{-1}o_i)$ denotes the continuous estimate of the gripper/object relative pose. Given the higher operating speed of the continuous pose estimator-controller (0.1$\sim$0.2 seconds per iteration) compared to the discrete pose estimator (2$\sim$6 seconds per iteration), the discrete pose estimation factor is integrated when an update is available every few steps within the continuous estimator-controller.

Similar to the discrete pose estimator, the environment (contact) prior is established during the initial time step:
\begin{align}
    &F_{cp}(c_1;c^*) = c^{*-1} c_1,
\end{align}
where $c^* \in SE(2)$ contains the prior information about the environment's orientation and its height.

Additionally, physics priors are imposed: %and kinematic constraints between the object and contact are introduced. 
\begin{align}
    &F_{vp}(\mathcal{V};\mathcal{V^*}) = \mathcal{V} - \mathcal{V^*}.
\end{align}
where $\mathcal{V^*}$ is the prior for the physics parameters.

Furthermore, a friction model rooted in the limit-surface model \cite{goyal1989planar,chavan2018hand} is imposed ($F_{fric}$). This model provides a correlation between the kinetic friction wrench and the direction of sliding at the grasp. In essence, it serves as a guide for predicting how the object will slide in response to a given gripper motion and extrinsic contact location. The correlation is formally represented as follows:
\begin{align}
    [\omega, v_x, v_y] \propto [\frac{M}{M_{max}^2}, \frac{F_x}{F_{max}^2}, \frac{F_y}{F_{max}^2}]. \label{eq:limitsurface}
\end{align}
Here, $[\omega, v_x, v_y]$ denotes the relative object twist in the gripper's frame, i.e. sliding direction, while $[M, F_x, F_y]$ signifies the friction wrench at the grasp. To fully capture the friction dynamics, additional kinematic and mechanical constraints at the extrinsic contact are also considered. These constraints are formulated as follows:
\begin{gather}
    M\hat{z} - \vec{l}_{gc} \times \vec{F} = 0, \label{eq:torquebal}\\%  \text{  (zero torque at the extrinsic contact)},\\
    v_{c, N}(g_{i-1},o_{i-1},c_{i-1},g_i,o_i) = 0, \label{eq:constnormal}\\
    v_{c, T}(g_{i-1},o_{i-1},c_{i-1},g_i,o_i) = 0 \qquad \label{eq:consttangent}\\ \qquad \perp (F_T = -\mu_{max} F_N \ \text{OR} \ F_T = \mu_{max} F_N ), \label{eq:frictioncone}
\end{gather}
In these equations, $\vec{l}_{gc}$  is the vector from the gripper to the contact point, and $v_{c, N}$ and $v_{c, T}$ represent the local velocities of the object at the point of contact in the directions that are normal and tangential to the environment, respectively. $F_N$ and $F_T$ denote the normal and tangential components of the force. Eq. \ref{eq:torquebal}  specifies that no net torque should be present at the point of extrinsic contact since we are assuming point contact. Eq. \ref{eq:constnormal} dictates that the normal component of the local velocity at the point of extrinsic contact must be zero as long as contact is maintained. Eq. \ref{eq:consttangent} and Eq. \ref{eq:frictioncone} work complementarily to stipulate that the tangential component of the local velocity at the contact point must be zero (Eq. \ref{eq:consttangent}), except in cases where the contact is sliding. In such instances, the contact force must lie on the boundary of the friction cone (Eq. \ref{eq:frictioncone}). By combining Eq. \ref{eq:limitsurface}$\sim$\ref{eq:frictioncone}, we establish a fully determined forward model for the contact and object poses, which allows the object pose at step $i$ to be expressed as a function of its previous poses, the current gripper pose, and the physics parameters:
\begin{equation}
    o_i^* = f(g_{i-1},o_{i-1},c_{i-1},g_i,\mathcal{V})
\end{equation}
This relationship can thus be encapsulated as a friction factor:
\begin{align}
    &F_{fric}(g_{i-1},o_{i-1},c_{i-1},g_i,o_i,\mathcal{V}) = o_i^{*-1}o_i.
\end{align}

Together, the estimation component formulates a smooth object pose trajectory that takes into account tactile measurements, robot kinematics, and physics model.

\begin{table*}[h]
\caption{Estimation Performance Comparison}
\centering
\begin{tblr}{hlines}
        & AUX                         & Pin                           & Stud                         & USB                          & Overall                     \\
{Single Shot\\(Tac2Pose \cite{bauza2022tac2pose})}      & {1.02\\(5.0 mm / 52.8 deg)} & {2.47\\(10.3 mm / 131.6 deg)} & {1.45\\(13.6 mm / 67.3 deg)} & {1.14\\(6.85 mm / 57.0~deg)} & {\textbf{1.52}\\(8.3 mm / 78.0 deg)} \\
{Discrete Est.\\(this paper)} & {0.17\\(2.8 mm / 6.8 deg)}  & {0.22\\(4.12 mm / 8.7 deg)}   & {0.20\\(4.6 mm / 5.9 deg)}   & {0.20\\(4.8 mm / 5.6 deg)}   & {\textbf{0.20}\\(3.9 mm / 6.9 deg)}  \\
{Continuous Est.\\(this paper)} & {0.09\\(1.7 mm / 3.2 deg)}  & {0.10\\(1.9 mm / 3.9 deg)}    & {0.08\\(2.2 mm / 2.1 deg)}   & {0.11\\(2.5 mm / 3.4 deg)}   & {\textbf{0.10}\\(2.0 mm / 3.3 deg)}  
\end{tblr}
\label{table:estimation}
\vspace{-4mm}
\end{table*}

\myparagraph{Control.} The control segment incorporates multiple auxiliary factors to facilitate the specification of regrasping objectives. First, the desired goal configuration is imposed at the end of the control horizon ($F_{goal}$). This comprises three distinct sub-factors, corresponding to the three manipulation objectives described in Section \ref{formulation}, which can be turned on or off, depending on the desired configuration:
\begin{enumerate}
    \item \( F_{\text{goal,go}} \) regulates the desired gripper/object relative pose at \( o_{t+T} \) and \( g_{t+T} \).
    \item \( F_{\text{goal,o}} \) enforces the object's orientation within the world frame at \( o_{t+T} \).
    \item \( F_{\text{goal,c}} \) dictates the desired contact point at \( c_{t+T} \), thereby facilitating controlled sliding interactions with the environment.
\end{enumerate}
These sub-factors are mathematically expressed as follows:
\begin{align}
    F_{goal,go}(g_{t+T},o_{t+T}) &= p_{o,goal}^{g \ -1} (g_{t+T}^{-1} o_{t+T}),\\
    F_{goal,o}(o_{t+T}) &= o_{goal}^{-1} o_{t+T},\\
    F_{goal,c}(c_{t+T}) &= c_{goal}^{-1} c_{t+T}.
\end{align}
Here, $p_{o,goal}^{g}$ signifies the target relative gripper/object pose, \( o_{goal} \) represents the desired object orientation in the world frame, and \( c_{goal} \) is the intended contact point.

Additionally, the \( F_{motion} \) factor minimizes the gripper motion across consecutive time steps, encouraging motion smoothness. Concurrently, a contact maintenance factor, \( F_{cm} \), serves as a soft constraint to direct the gripper's motion in a way that prevents it from losing contact with the environment:
\begin{align}
    F_{cm}(g_{i-1}, &c_{i-1}, g_i; \epsilon_i) = max(0, \zeta_i(g_{i-1}, c_{i-1}, g_i) + \epsilon_i),
    %&= max\left(0, \left(c_{i-1}^{-1} (g_{i-1}^{-1} g_{i} g_{i-1}^{-1} c_{i-1})\right)_y - \epsilon_i \right).
\end{align}
\( \zeta_i \) represents the normal component of the virtual local displacement from step  \( i-1 \) to \( i \) at the contact point. The term \( \epsilon_i \) is a small positive scalar, encouraging \( \zeta_i \) to be negative, thus fostering a motion that pushes against the environment.

Taken together, these factors cohesively formulate a motion plan, which is then communicated to the robot. The robot continues to follow the interpolated trajectory of this motion plan until it receives the next update.

\section{Experiments and Results}

We conducted a series of experiments on four distinct 3D-printed objects (illustrated in Fig. \ref{fig:objects}) to validate the efficacy of our algorithm. The experiments were designed to:
\begin{enumerate}
\item Evaluate the algorithm's performance across a variety of target configurations.
\item Assess the algorithm's applicability to specific real-world tasks, such as object insertion.
\end{enumerate}

\subsection{Performance Across Various Goal Configurations} \label{subsection:varigoal}

We assessed our algorithm's performance using a total of 18 diverse goal configurations. Our framework allows for specifying goals relative to the gripper (regrasping) and relative to the world frame (reorienting), facilitating different downstream tasks. For example, regrasping can improve grasp stability, enable tactile exploration, and establish a grasp optimized for both force execution and the avoidance of collisions or kinematic singularities in downstream tasks. On the other hand, reorienting the object can enable mating with target objects in the environment or prevent collisions with obstacles. The configurations we evaluate fall into four distinct categories:

\begin{itemize}
    \item Relative Orientation + Stationary Extrinsic Contact
    \item Relative Orientation/Translation + Stationary Extrinsic Contact
    \item Relative Orientation + Global Orientation + Stationary Extrinsic Contact
    \item Relative Orientation + Sliding Extrinsic Contact
\end{itemize}

Examples of these four goal configuration types are illustrated in Fig. \ref{fig:samples}, along with corresponding plots showcasing estimation accuracy. The red silhouettes that move along with the gripper represent the desired relative pose between the gripper and the object. Conversely, the grey silhouettes depict object poses as measured by Apriltags, which we use as the ground truth object pose. The red dots mark the desired extrinsic contact location. In Fig. \ref{fig:samples}c, the second red silhouette signifies the desired object orientation in the global frame. The time series plots in the right column depict the normalized estimation error of the proposed model ('Continuous'), alongside two ablation models: 1) single-shot Tac2Pose estimation ('Single Shot'), 2) discrete filtered estimation from Viterbi decoding ('Discrete'). These results attest to the algorithm's adeptness in attaining desired goal configurations while showing better estimation performance compared to the ablation models.

A summary of each algorithm's average estimation performance is presented in Table \ref{table:estimation}. The values denote the normalized estimation error, computed as follows:
\begin{equation}
    \epsilon_{\text{norm}} = ||(\epsilon_{\text{rot}}, \epsilon_{\text{trn}} / (l_\text{obj}/2)||_1
\end{equation}
Here, $||\cdot||_1$ signifies the L1-norm, $\epsilon_{\text{rot}}$ indicates rotation error in radians, $\epsilon_{\text{trn}}$ denotes translation error, and $l_\text{obj}$ represents the object's length. This analysis reveals a marked reduction in overall normalized error, progressing from 1.52 to 0.20 when transitioning from single-shot estimation to the discrete pose estimator's filtered estimation. The single-shot estimator suffers due to ambiguity in individual tactile images, as explored thoroughly in \cite{bauza2022tac2pose}. For most grasps of the objects we experiment with, a single tactile imprint is not sufficient to uniquely localize the object. The discrete pose estimator is able to reduce ambiguity by fusing information over a sequence of tactile images, obtained by traversing the object surface and therefore exposing the estimator to a more complete view of the object geometry. The discrete pose estimator also consumes information about the ground height and orientation, providing an additional constraint on the object pose. In this way, the tasks of tactile object pose estimation and in-hand manipulation are synergistic: tactile object pose estimation supervises in-hand manipulation, while in-hand manipulation allows object pose to be unambiguously estimated from tactile images. Furthermore, a subsequent improvement is observed, decreasing from 0.20 to 0.10 when employing the continuous pose estimator-controller. This improvement comes from refining the estimation accuracy beyond the resolution of the discrete grid.

\subsection{Real-World Application: Insertion Task}

To validate our algorithm's practical utility, we applied it to a specific downstream task — object insertion with small clearance (1$\sim$0.5 mm). For these experiments, we sampled random goal configurations from the first category (adjusting relative orientation) described in Section \ref{subsection:varigoal}. Following this, we aimed to insert the grasped object into holes with 1 mm and 0.5 mm total clearance in diameter.

Table \ref{table:insertion} summarizes the outcomes of these insertion attempts. The AUX connector, which features a tapered profile at the tip, had a success rate exceeding 90\%. On the other hand, the success rate dropped considerably for objects with untapered profiles, especially when the clearance was narrowed from 1 mm to 0.5 mm. The varying performance is consistent with our expectations, given that the algorithm’s average pose estimation accuracy is approximately 2.0 mm as quantified in Section \ref{subsection:varigoal}.

These findings indicate that our algorithm is useful in tasks that necessitate regrasping and reorienting objects to fulfill downstream objectives by meeting the goal configuration. However, for applications requiring sub-millimeter accuracy, the algorithm's performance would benefit from integration with a compliant controlled insertion policy (e.g., \cite{inoue2017deep, dong2021tactile, zhao2022offline, kim2022active}).

\begin{table}[h]
\caption{Insertion Experiment Results (Success/Attempt)}
\centering
\begin{tabular}{|c||c|c|c|c|} 
\hline
Clearance & AUX     & Pin    & Stud   & USB     \\ 
\hline
1 mm   & 10 / 10 & 6 / 10 & 7 / 10 & 7 / 10  \\ 
\hline
0.5 mm & 9 / 10  & 3 / 10 & 5 / 10 & 6 / 10  \\
\hline
\end{tabular}
\label{table:insertion}
\vspace{-4mm}
\end{table}

\section{Conclusions}

This paper introduces a simultaneous tactile estimator-controller tailored for in-hand object manipulation. The framework harnesses extrinsic dexterity to regrasp a grasped object while simultaneously estimating object poses. This innovation holds particular promise in scenarios necessitating object or grasp reorientation for subsequent tasks like insertion, particularly in cases where precise visual perception of the object's global pose is difficult due to occlusions. In future research, we aim to explore methodologies for autonomously determining optimal target orientations for task execution, rather than relying on manual specification.

%\addtolength{\textheight}{-12cm}

\bibliographystyle{IEEEtran}
\bibliography{ref}

\end{document}